# A Unified Approach for Modeling and Recognition of Individual Actions and Group Activities

Qiang Qiu, Rama Chellappa

Department of Computer Science, University of Maryland, College Park
qiu@cs.umd.edu, rama@umiacs.umd.edu

**Abstract.** Recognizing group activities is challenging due to the difficulties in isolating individual entities, finding the respective roles played by the individuals and representing the complex interactions among the participants. Individual actions and group activities in videos can be represented in a common framework as they share the following common feature: both are composed of a set of low-level features describing motions, e.g., optical flow for each pixel or a trajectory for each feature point, according to a set of composition constraints in both temporal and spatial dimensions. In this paper, we present a unified model to assess the similarity between two given individual or group activities. Our approach avoids explicit extraction of individual actors, identifying and representing the inter-person interactions. With the proposed approach, retrieval from a video database can be performed through Query-by-Example; and activities can be recognized by querying videos containing known activities. The suggested video matching process can be performed in an unsupervised manner. We demonstrate the performance of our approach by recognizing a set of human actions and football plays.

## 1 Introduction

Modeling and recognition of group activities can generally be performed in two ways. In the first category, individuals are isolated to decompose the problem into multiple single-person scenarios, which can be handled with many existing single-person action models. This approach will fail due to occlusion of body parts and not incorporating inter-person interaction. In the second category, person interactions are encoded into structures like Bayes Networks [1], [2], [3], Petri nets [4], stochastic grammars [5], etc. This approach is successful in modeling predetermined, structured multi-person activities. Due to the complexity and unpredictability of human interactions, many real-world activities do not obey the encoded interaction constraints. In both categories, it is hard to tell the *entity correspondences*, i.e., matchings between entities in the model and entities in the observation. When the number of persons involved in an activity is large, e.g., 22 players in a football match, the problem is often simplified with an assumption that the entity correspondences are known [6] or easy to obtain [1].

In this paper, we propose a simple matching method to recognize single or multi-person actions or retrieve actions similar to the given one. This method avoids the strong known entity correspondences assumption, unreliable human body extraction and modeling of complex human interactions. The key characteristics of the proposed approach can be summarized as follows,

1. It handles both single and multiple person actions in a unified way. Thus, we can apply it regardless of the number of persons in a video.
2. It does not assume that the correspondence between entities detected in videos and entities specified in a model are known, which is a strong assumption made in many existing group activity recognition methods.
3. It avoids explicit extraction of human body structures and encoding of complex human interactions.
4. It is insensitive to action executing rate/order, occlusion and viewing angles.
5. It supports unsupervised learning.

The rest of this paper is structured as follows. In Sec. 2, we discuss a unified way to describe both individual actions and group activities using Markov Logic Networks, and suggest a trajectory-based activity representation. Then, in Sec. 3, we construct from such a unified representation a general probabilistic graphical model to assess the semantic similarity between an observed action/activity to a given model. Experimental results are given in Sec. 4 to demonstrate the effectiveness of our approach in handling actions and activities in a unified way.

### 1.1 Previous Work

Extensive research has been conducted for modeling and recognition of human activities. Most existing work has focused on modeling and recognition of single person actions, including early approaches like 2D-templates model [7], hidden Markov model [8], or more recent approaches based on bag-of-words model [9], linear dynamical systems [10], etc.. Most of these approaches can not be directly applied to group activities due to the inherent difficulties in modeling inter-person interactions.

Group activities have been mostly modeled using Belief Networks [1], [2], [3], [11], or other types of models like Petri nets [4]. Though many of these approaches are successful in modeling various group activity scenarios, they suffer from the following drawbacks. 1.) Manual specification of model structures is often required [4], [1]. Given the complex and unpredictable nature of human interactions, it is difficult to manually specify a comprehensive activity model. 2.) Models are often designed to handle specific types of activities, e.g., football plays [1], pairwise activities [11], etc. It can be difficult for extensions to activities involving more persons or other scenarios. 3.) Techniques for matching entities in a video and entities in the model are often not carefully addressed [6], [1]. Given activities like football plays, which involve 22 players, such entity correspondence problem can not be trivially handled.

## 2 Unified Action and activity Representation

Human activities can be briefly classified as individual actions and group activities. In this paper, thereafter, we refer to individual motion patterns as **actions**, and coordinated actions involving multiple humans as **activities**. We discuss here a unified way to describe both actions and activities based on the Markov Logic Networks framework (MLN [12]).

### 2.1 Activity Representation using Markov Logic Networks

**Markov Logic Networks** Markov Logic Networks are graphical models developed to combine logical and probabilistic reasoning. An MLN is a set of pairs $(F_i, w_i)$ where each first-order logic formula $F_i$ is associated with a non-negative real value weight $w_i$. Grounding is to replace all variables in a term by constants. An MLN is a template to ground a Markov Random Field (MRF) as follows:

- Each ground predicate in MLN is mapped to one node in MRF.
- Each ground logic formula $F_i$ in MLN corresponds to a clique in MRF over involved ground predicates, with weight $w_i$ and a feature $f_i$ of value 1 if the formula is true or 0 otherwise.

In such an MRF, the probability of a possible world is proportional to $\exp(\sum_i (w_i f_i))$.

In [13], the MLN has been used to model parking lot visual events through a set of identified entities, i.e., cars (C), humans (H) and locations (L); and a vocabulary consisting of predicates like *carLeave(C), openTrunk(C,H), enter(C,H), disappear(H,L)*, etc. Different predicates are modeled and recognized separately through object detection and tracking techniques. For example, *openTrunk(C,H)* is evaluated through motion patterns in the trunk area of a car.

**Challenges to Model Activity in MLN** Given the expressive power of MLN in modeling complex visual events, we still face the following challenges to obtain actual discriminative models for activity recognition,

- It is difficult to associate entities in a video with entities specified in an activity model. For example, from a video sequence, it is hard to map each of 22 players in a football play shown in Fig. 1 to the corresponding roles specified in the model. This issue causes many commonly used discriminative models to fail due to the inconsistent feature orderings across different datasets.
- Matching a given video to an activity model through a simple search is computationally intensive. For example, it takes $\frac{N!}{(M-N)!}$ matches, e.g., 22! in football scenarios, for entity correspondence, where $M$ and $N$ are the number of entities in the observation and model respectively.
- An MLN activity model is domain specific.
- Often only part of an activity is observed. Possible reasons for such incomplete observations can be many: entities are occluded or excluded from scenes, or just part of an activity is captured in the video. It is challenging to model missing data in MLN.

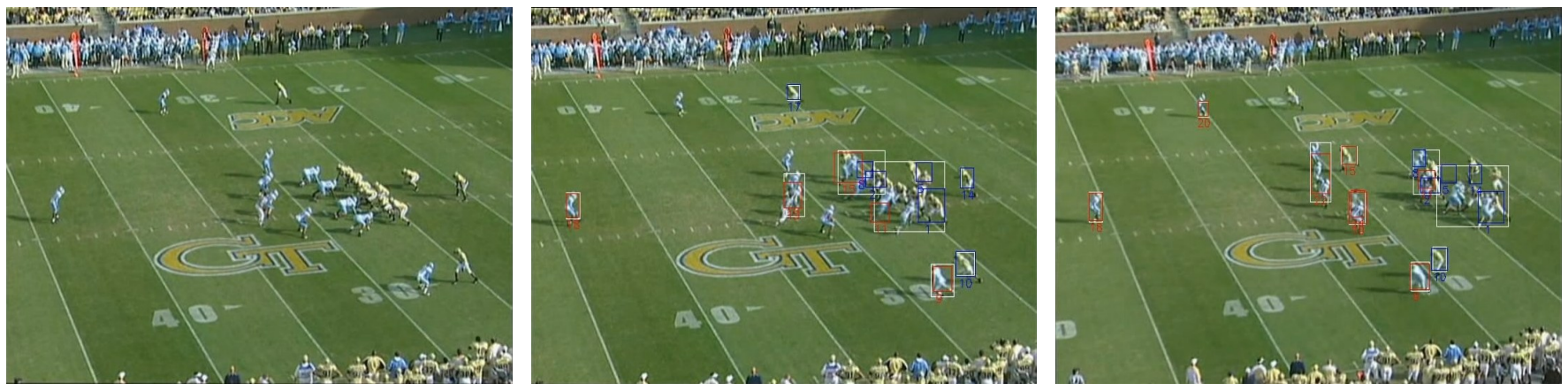

Fig. 1: Sample frames of a football *Hitch* play video sequence

**Activity Modeling in MLN** To address the challenges above, we start with an approach similar to [13] which describes human actions or activities using entities and predicates. For example, the *running* action is interpreted as $\exists$ $b_1$, $b_2$, *LeftLeg*($b_1$) $\wedge$ *RightLeg*($b_2$) $\wedge$ *RunMotion*($b_1$) $\wedge$ *RunMotion*($b_2$) $\wedge$ *Simultaneous*(*RunMotion*($b_1$), *RunMotion*($b_2$)). A *Simple-p51curl* football play activity involving 4 offensive players and 1 defensive player in Fig. 2a, which can be sketched by a coach, is represented in Fig. 2b, where we list the temporal constraints among motions. One can also incorporate spatial constraints such as orientation and distance.

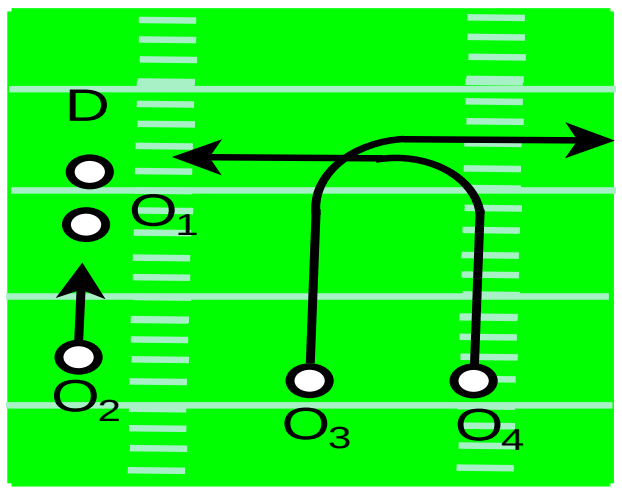

(a) Chalkboard sketch [1]

$\exists\, p_1, p_2, p_3, p_4, p_5,$
Offensive($p_1$) $\wedge$ Offensive($p_2$) $\wedge$ Offensive($p_3$)
$\wedge$ Offensive($p_4$) $\wedge$ Defensive($p_5$) $\wedge$ GoStraightUp($p_3$)
$\wedge$ TurnRight($p_3$) $\wedge$ GoStraightUp($p_4$) $\wedge$ TurnLeft($p_4$)
$\wedge$ GoStraightUp($p_2$) $\wedge$ StandStill($p_1$) $\wedge$ StandStill($p_5$)
$\wedge$ Sequential (GoStraightUp($p_3$), TurnRight($p_3$))
$\wedge$ Sequential (GoStraightUp($p_4$), TurnLeft($p_4$))
$\wedge$ Simultaneous(GoStraightUp($p_3$), GoStraightUp($p_4$))
$\wedge$ Simultaneous(TurnRight($p_3$), TurnLeft($p_4$))
$\wedge$ Simultaneous(GoStraightUp($p_3$), GoStraightUp($p_2$))
$\wedge$ Simultaneous(GoStraightUp($p_2$), StandStill($p_1$))
$\wedge$ Simultaneous(StandStill($p_1$), StandStill($p_5$))

(b) Semantic interpretation

Fig. 2: The football *simple-p51curl* play

From the above two examples, one can notice that a formula for each action or activity consists of three types of predicates describing respectively ***entities***, e.g., *Offensive*($p_3$), ***atomic motions***, e.g., *TurnRight*($p_3$), and ***pairwise motion constraints***, e.g., *Simultaneous*(*TurnRight*($p_3$), *TurnLeft*($p_4$)). It is noted that an entity here is defined as any moving person, body part, or object. A group of people moving in a coordinated way sometimes can be as a whole considered as an entity. Though only we limit ourselves to pairwise interactions, higher order relationships can be introduced using more complex models.

We now address the following two specific problems to make MLN more appropriate for activity modeling.

***1.*** *An MRF constructed from an MLN that models activities can easily contains a large number of nodes.* During grounding, an existential quantifier in MLN is expended over the entire entity domain to obtain a disjunction of the original formula. For example, an activity is given as a formula $\exists p_1, p_2$, *TurnRight*($p_1$)$\wedge$ *TurnLeft*($p_2$), and two entities $P_1$ and $P_2$ are detected in the video. Since it is typically difficult to associate entities in a video with entities specified in the formula, the grounded formula will be (*TurnRight*($P_1$)$\wedge$ *TurnLeft*($P_2$)) $\vee$ (*TurnRight*($P_2$)$\wedge$ *TurnLeft*($P_1$)). Thus, using the MLN exhaustive grounding scheme, we can ground extremely complex MRF structure from an MLN, e.g., a network corresponds to the disjunction of 22! conjunctive clauses for a 22 player football activity. To avoid such a situation, we suggest to ground first the MRF from an activity formula as if we knew the correspondence of detected entities in video with the ones specified in the model, and postpone the discovery of the optimal grounding scheme until the inference stage, which is described later in Sec. 3. So the MRF is constructed by simply *mapping each atomic motion to a node and each pairwise constraint to an edge*. Such a network can also be easily described in terms of cliques or factors.

***2.*** *Each predicate can require separate manual modeling to encode the knowledge* [13], which is a tedious task. To avoid such predicate manual modeling, we suggest direct representation of each atomic motion as a set of low-level vision features, e.g., optical flows for a set of pixels or trajectories for a set of feature points, and describing pairwise constraints among atomic motions as temporal and spatial measures that can be obtained directly from videos. One such predicate modeling that requires no manual efforts is suggested in Sec. 2.2.

It is noted that, we limit an activity formula as the conjunction form of extensional predicates, i.e., their values can be directly evaluated by the low level vision features. The detection probability of each extensional predicate under an optimal grounding scheme will provide the corresponding weight in MLN. Given an optimal grounding scheme and evaluated extensional predicates, intensional predicates, whose values can only be inferred, and other logical operations can be added by following the discussions in [13].

In the rest of this section, we will describe in details how to represent motions using point trajectories as features. With this representation scheme, both individual actions and group activities can be interpreted as a collection of atomic motions of a set of landmark points and a set of pairwise constraints between every pair of point atomic motions. In Sec. 3, we will give a formal treatment of embedding the discovery of optimal ground scheme in the inference process.

## 2.2 Trajectory-based Activity Representation

Trajectory is a simple and direct description of motion landmarks. Motion trajectories can be obtained from a video through tracking algorithms, and represented as a sequence of $\langle$ *x(t), y(t)* $\rangle$, i.e., the $x$ and $y$ coordinates of a landmark point at time $t$. In this paper, we adopt point trajectories as features representing motions in a video to automatically generate models for activity predicates.

**Joint Segmentation of Multiple Trajectories for Atomic Motions** Multiple landmark points are often chosen for each entity, thus, the motion of an entity is described by a group of trajectories. To avoid manual modeling of each atomic motion predicate in MLN, we decompose the entity motion into a set of atomic motions by jointly segmenting the multiple corresponding trajectories based on the underlying *motion discontinuities*. Motion discontinuities can be reflected by the change in velocity and acceleration, which are measured as the first and second derivatives of the trajectory. We adopt the spatio-temporal curvature (1) defined in [14] as a measure of motion discontinuities.

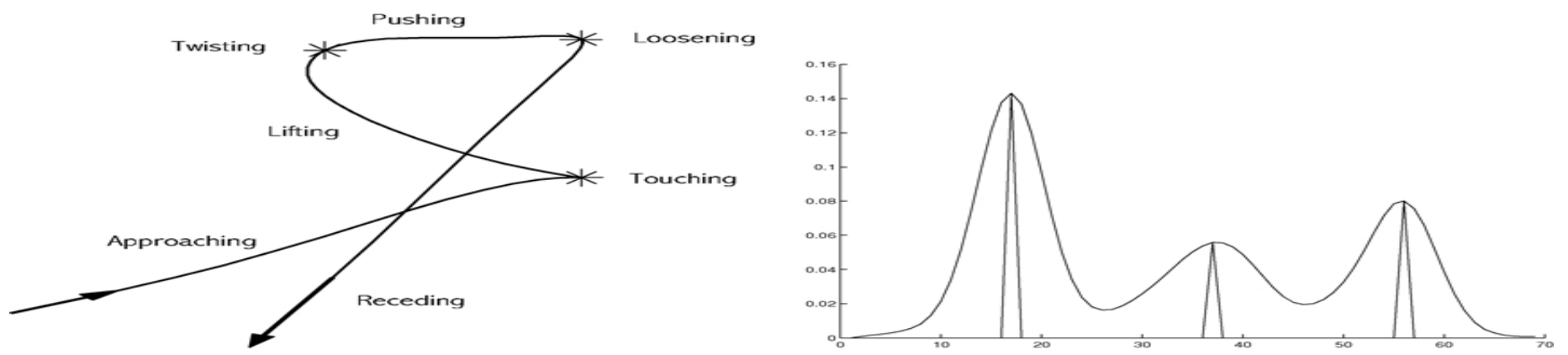

Fig. 3: Open cabinet trajectory of single hand landmark point and its curvature (taken from [14])

$$\kappa(t) = \frac{\sqrt{y''(t)^2 + x''(t)^2 + (x'(t)y''(t) - x''(t)y'(t))^2}}{(\sqrt{x'(t)^2 + y'(t)^2 + 1})^3} \tag{1}$$

As shown in Fig. 3, for a single trajectory, motion discontinuity instants correspond to local maxima of its spatio-temporal curvature. Thus, curvature local maxima suggest good segmentation points. Extensions are required to handle joint segmentation of multiple trajectories.

For individual actions, neighboring landmarks can share common motions due to human articulation constraints; for structured group activities, collaborative players can have correlated motions. We map the trajectories of all landmark points to the embedding space using manifold learning techniques, such as Laplacian eigenmaps [15]. We adopt the first $d$, e.g., $d = 3$, components of the embedding space as major motion components. After being grouped by major motion components through k-means clustering, trajectories to be jointly segmented, which are likely from the same entity, exhibit highly correlated motion discontinuities. By summing up the curvatures of each individual trajectory, the local maxima of the *aggregated curvature* indicate joint motion discontinuities. Validations of trajectory joint segmentation are shown in Sec. 4.1.

The spatio-temporal curvatures are view invariant as discussed in [14]. We further represent each segmented trajectory with a constant number of samples for not differentiating the time spent between two motion discontinuity instants. The resulting atomic motions are not affected by changes in view and rate.

**Pairwise Temporal Relationship** We now discuss pairwise motion constraints that can be obtained directly from videos. Given the durations of two atomic motions as $\langle s_1, e_1 \rangle$ and $\langle s_2, e_2 \rangle$, where $s_1, e_1, s_2, e_2$ specify the corresponding start and end frame numbers. Based on Allen's Interval Algebra [16], temporal relationships are derived as follows,

- ***equal*** $(s_1 = s_2) \wedge (e_1 = e_2)$,
- ***during*** $(s_1 < s_2) \wedge (e_1 > e_2)$,

– ***overlap*** $(s_1 < s_2) \wedge (e_1 > s_2) \wedge (e_1 < e_2)$,
– ***after*** $s_2 > e_1$
– ***before*** $s_1 > e_2$

Based on such a categorization, we adopt the following more compact form that enables efficient inferencing methods. We combine *equal, during, overlap* as *simultaneous* relationship, and suggest the quantitative measure for *simultaneous* relationship as $\max(|s_1 - s_2|, |e_1 - e_2|)$. We combine *before, after* as *sequential* relationship, and suggest the quantitative measure for *sequential* relationship as $\min(|e_1 - s_2|, |e_2 - s_1|)$. Based on the such quantitative measure scheme, we obtain a reasonable definition for neighborhood of two atomic motions required in Sec. 3.3 as, $\min(\max(|s_1 - s_2|, |e_1 - e_2|), \min(|e_1 - s_2|, |e_2 - s_1|))$.

**Pairwise Spatial Relationship** Given the centroids of two atomic motions as $m_i$ and $m_j$, the spatial distance between the locations of the two motions is measured by $||m_i - m_j||$. The spatial orientation relationship between two motions is measured by $\frac{m_i \cdot m_j}{||m_i||\ ||m_j||}$.

## 3 Probabilistic Semantic Matching Model

Our main objective is to, at the semantic level, assess how similar an observed event is to a specified event, when are both described in MLNs. For brevity, we use the term *event* for both individual action and group activity. As discussed in Sec. 2, we obtain from MLNs a grounded MRF $Z$ for the observed event and an MRF $Y$ for the specified event, by postponing until the inference stage the discovery of the actual entity correspondence between the observation and the model. Our basic approach is to find an optimal way to match nodes in $Z$ to those in $Y$. Such an optimal matching, denoted as $X$, will maximize the semantic similarity $p(Z|Y)$ defined in Sec. 3.2 and also imply an optimal grounding scheme for $Z$. Given the maximized similarities to multiple event models, an observed event can be classified. The process to assess similarity can easily become intangible. In Sec. 3.3, we derive a closed form variational approximation to semantic similarities, and also suggest a practical alternative for designing an efficient similarity inferencing engine with local dependency assumption.

### 3.1 Model Overview

We discuss here in details the event model, the observation model, and the semantic mapping shown in Fig. 4.

**Event Model Y** We construct an event model $Y$ from an MLN that specifies an event by mapping each atomic motion to a node and each pairwise constraint to an edge. Given $M$ atomic motions in the event specification, $Y$ is defined as,

$$Y = \{y_m, y^s_{i,j}, y^t_{i,j} : m, i, j \in [1, M]\}$$

where the node $y_m$ corresponds to an atomic motion; the edges $y^s_{i,j}$ and $y^t_{i,j}$ corresponds to the spatial and temporal constraints respectively between two atomic motions $y_i$ and $y_j$. Multiple edges between two nodes are eventually aggregated into one by adding up the associated potentials introduced later on. With Query-by-Example, $Y$ can be understood as the graphical representation for each given event to be queried, which are point trajectories in this paper.

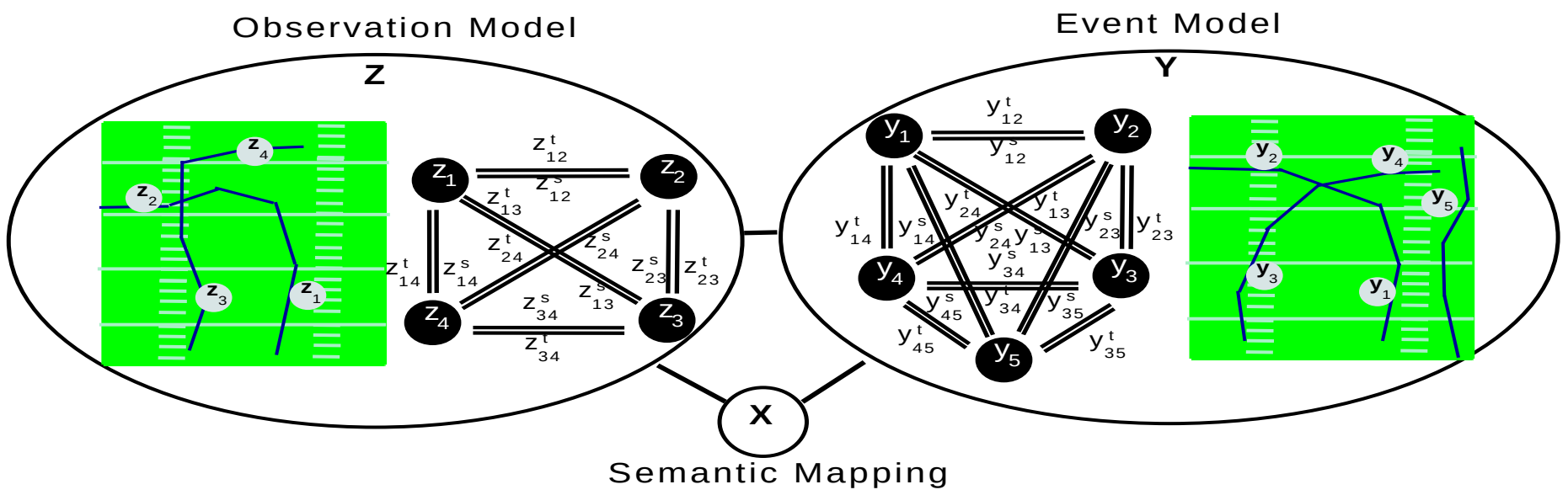

Fig. 4: Overview of the semantic matching model. $X$ denotes semantic mapping, $Y$ is an event model built for each event to be queried, and $Z$ represents a video observation. In both event model $Y$ and observation model $Z$, each node denotes an atomic motion, and each edge denotes a pairwise constraint between two atomic motions. $X$ maps nodes in $Z$ to nodes in $Y$.

**Observation Model Z** We construct the observation model $Z$ in the same way as $Y$. Considering $N$ atomic motions in an observed event, $Z$ is defined as

$$Z = \{z_n, z^s_{i,j}, z^t_{i,j} : n, i, j \in [1, N]\}$$

where the node $z_n$ corresponds to an observed atomic motion, the edges $z^s_{i,j}$ and $z^t_{i,j}$ corresponds to the spatial and temporal constraints respectively between two atomic motions $z_i$ and $z_j$. Using Query-by-Example, $Z$ is the graphical representation for an event used to pose a query.

**Semantic Mapping X** Our objective is to find an optimal way to match nodes in an observation model $Z$ to nodes in an event model $Y$. Given $M$ nodes in $Y$ and $N$ nodes in $Z$, each observed node $z_i$ is associated with a mapping variable $x_i$ of a discrete value range $[1, M]$. For example, assigning a value $j$ to $x_i$ indicates that the observed motion $z_i$ is mapped to the motion $y_j$ specified in the event model. The matching process is to find optimal values for each $x_i \in X$.

### 3.2 Probabilistic Modeling

We formally describe the semantic matching model using Markov Random Fields. The conditional likelihood is modeled as $p(Z|X,Y) = \frac{1}{C_l} e^{-U(Z|X,Y)}$ , where $C_l$ is a normalization constant. Given $N$ nodes in $Z$, the likelihood energy $U(Z|X,Y)$ is defined as the sum of clique potentials,

$$U(Z|X,Y) = \sum_{i \in [1,N]} V_1(z_i|x_i, Y) + \sum_{i,j \in [1,N], j>i} [V_2(z^s_{i,j}|x_i, x_j, Y) + V_2(z^t_{i,j}|x_i, x_j, Y)]$$

where $V_1(z_i|x_i, Y)$ is the node potential, and, to facilitate the discussion, we denote the spatial and temporal pairwise potentials separately as $V_2(z^s_{i,j}|x_i, x_j, Y)$ and $V_2(z^t_{i,j}|x_i, x_j, Y)$. The prior distribution is $p(X|Y) = \frac{1}{C_p}e^{-U(X|Y)}$ , where $C_p$ is a normalization constant and the prior energy is defined as $U(X|Y) = \sum_{i\in[1,N]} V_1(x_i|Y) + \sum_{i,j\in[1,N],j>i} V_2(x_i, x_j|Y)$.

**Node Potential** Each node $z_i$ in the observation model $Z$ represents an observed atomic motion consisting of $P$ trajectories, given $P$ landmark points associated with the observed entity. We use $y_{x_i}$ to denote the node in the event model $Y$ indicated by the mapping value $x_i$. In query-by-example paradigms, $y_{x_i}$ represents an atomic motion consists of $Q$ trajectories, given $Q$ points with the specified entity. By assuming zero-mean Gaussian observation noise, $y_{x_i}$ specifies $Q$ number of Gaussian distributions with its $q^{th}$ trajectory $y^{(q)}_{x_i}$ as the mean for the $q^{th}$ distribution. We assume the same noise deviation $\sigma_{x_i}$ for distributions in $y_{x_i}$. In the supervised mode, $\sigma_{x_i}$ can be learned from training data; in the unsupervised mode, we estimate $\sigma_{x_i}$ from observations $Z$. Let $z^{(p)}_i$ denote the $p^{th}$ trajectory in node $z_i$, the node energy is obtained by maximizing

$$V_1(z_i|x_i, Y) = \sum_{\{p,q\}\in S} \frac{(z^{(p)}_i - y^{(q)}_{x_i})^T (z^{(p)}_i - y^{(q)}_{x_i})}{2\sigma^2_{x_i}}$$

with a set $S$ which indicates the best trajectory correspondence between $z_i$ and $y_{x_i}$. As each atomic motion consists of a small number of short trajectory segments, $S$ can be easily obtained.

**Pairwise Potential** We use $z^t_{i,j}$ to denote the temporal relationship between two observed atomic motions $z_i$ and $z_j$, and $y^t_{x_i,x_j}$ as the temporal relationship between two atomic motions indicated by the mapping value $x_i, x_j$ in the event model. We adopt quantitative measures for motion relationships in Sec. 2.2. With zero-mean Gaussian noise, the temporal pairwise potential is written as,

$$V_2(z^t_{i,j}|x_i, x_j, Y) = \frac{(z^t_{i,j} - y^t_{x_i,x_j})^T (z^t_{i,j} - y^t_{x_i,x_j})}{2{\sigma^t_{x_i,x_j}}^2}$$

where $\sigma^t_{x_i,x_j}$ is the noise deviation for temporal relationship observation and is handled similarly as node observation noise deviation. Spatial pairwise potential $V_2(z^s_{ij}|x_i, x_j, Y)$ is obtained in the same way.

**Semantic Similarity Measure** Given the likelihood distribution $p(Z|X, Y)$ and the prior distribution $p(X|Y)$, the *semantic similarity* between an observed event $Z$ and an event model $Y$ is defined as,

$$p(Z|Y) = \sum_{X\in\mathcal{X}} p(Z, X|Y) = \sum_{X\in\mathcal{X}} \frac{1}{C_l C_p} e^{-U(Z,X|Y)}, \quad where \;\; U(Z, X|Y) = \tag{2}$$

$$\sum_{i\in[1,N]} [V_1(z_i|x_i, Y) + V_1(x_i|Y)] + \sum_{i,j\in[1,N],j>i} [V_2(z^s_{i,j}|x_i, x_j, Y) + V_2(z^t_{i,j}|x_i, x_j, Y) + V_2(x_i, x_j|Y)]$$

The optimality of a semantic matching is assessed by $p(X|Y, Z) = \frac{p(X,Z|Y)}{p(Z|Y)}$.

### 3.3 Semantic Similarity Approximation

It is not difficult to notice that the event similarity in (2) requires a marginalization over $\chi$, the possible domain of $X$, which is of size $O(N^M)$. Here $N$ and $M$ are the node numbers in the observation and the event model respectively. *Such marginalization step can easily become intangible.* In this section, we first derive a closed form variational approximation to the event similarity. In [17], good discussions can be found on how to construct a variational approximation. We then suggest an alternative for inexact inference by assuming local dependence.

**Similarity Variational Approximation** Our variational formalism is the process to derive a distribution $q(X)$ which approximates the true posterior $p(X|Y,Z)$. By introducing an arbitrary distribution $q(X)$, the lower bound on the logarithm of the event similarity, $\log p(Z|Y)$, can be obtained through Jensen's Inequality, $\log \sum_x f(x) \geq \sum_x \log f(x)$, as follows,

$$\begin{aligned}\log p(Z|Y) = \log \sum_{X\in\chi} q(X)\frac{p(Z,X|Y)}{q(X)} &\geq \sum_{X\in\chi} q(X)\log p(Z|Y) - KL(q(X)||p(X|Y,Z)) \\ &= \log p(Z|Y) - KL(q(X)||p(X|Y,Z)) \qquad (3)\end{aligned}$$

We use $LS(q)$ to denote the lower bound of $\log p(Z|Y)$, i.e., the right side of (3). By the non-negative property of Kullback-Leibler (KL) distance, we can maximize $LS(q)$ by minimizing the KL-distance between $q(X)$ and $p(X|Y,Z)$. When $q(X)$ equals $p(X|Y,Z)$, the above lower bound becomes equality.

In variational inference, one typical approximation step is to assume independence among hidden variables [17]. In our case, we need to assume that the semantic mappings $X$ are independent of each other to simplify the approximation process, i.e., $q(X) = \prod_{i=1}^{N} q(x_i)$. Though such independence assumption will affect the ultimate approximation accuracy, as pointed out in [17], we still have sufficient degrees of freedom, $(M-1)N$, for non-trivial approximated $p(X|Y,Z)$, where $|Y| = M$ and $|Z| = N$.

With the above formulation, we adopt an E-M type methodology such that, in each iteration, we fix a set of parameters and find the marginal $q(x_k)$ to maximize $LS(q)$ through $\frac{\partial}{\partial q(x_k)} LS(q) = 0$. Based on (2), we define the following abbreviated expressions,

$$\begin{aligned}V_1(i) &= V_1(z_i|x_i,Y) + V_1(x_i|Y) \\ V_2(i,j) &= V_2(z^s_{i,j}|x_i,x_j,Y) + V_2(z^s_{i,j}|x_i,x_j,Y) + V_2(x_i,x_j|Y)\end{aligned}$$

and rewrite $LS(q)$ in terms of $x_k$ and add the constraint that $\sum_{x_k} q(x_k) = 1$,

$$\begin{aligned}LS(q) = &- \sum_{i\in[1,N]\setminus\{k\}} \sum_{x_i} q(x_i)[\log q(x_i) + V_1(i)] - \log C_l C_p - \lambda(\sum_{x_k} q(x_k) - 1) \qquad (4) \\ &- \sum_{x_k} q(x_k)[\log q(x_k) + V_1(k)] - \sum_{x_k} q(x_k) \sum_{i,j\in[1,N]\setminus\{k\},j>i} \sum_{x_i,x_j} q(x_i)q(x_j)V_2(i,j)\end{aligned}$$

where $q(x_i)$ is the marginal probability over $x_i$, and $\lambda$ is a Lagrange parameter. Now, by omitting those constants wrt. $x_k$ in (4), $\frac{\partial}{\partial q(x_k)} LS(q) = 0$ yields,

$$q(x_k) = \frac{1}{C_q} e^{-\{\sum_{i,j\in[1,N]\setminus\{k\},j>i} \sum_{x_i,x_j} q(x_i)q(x_j)U_2(i,j)\} - U_1(k)}$$

where $C_p$ is the normalization constant. The derived equation will be used for each update of the optimal $q(x)$. We can then use $q(X)$ as an approximation to $p(X|Y,Z)$, and $e^{LS(q)}$ as an approximation to $p(Z|Y)$.

**Similarity Approximation with Local Dependence** By assuming local dependence, i.e., the semantic mapping for a node will only depend on its neighbors, we can also adopt the iterative conditional mode (ICM) methodology [18] to efficiently compare the similarities between an observation and multiple models. It is critical to define node neighborhood in ICM. In Sec. 2.2, we define the neighborhood for atomic motions based on their temporal distance. We start with some initial semantic mapping configuration. From the following equations,

$$p(X|Y,Z) = p(x_k|X_{\setminus\{x_k\}}, Y, Z)p(X_{\setminus\{x_k\}}|Y,Z)$$

where $X_{\setminus\{x_k\}}$ indicates the set of variables $X$ except $x_k$. We notice that, to obtain the optimal mapping $X^*$, in each iteration we can update $x_k$ to maximize $p(x_k|X_{\setminus\{x_k\}}, Y, Z)$, since $p(X|Y,Z)$ will never decrease with such a scheme, leading to eventual convergence. By the local dependence assumption, we can rewrite $p(x_k|X_{\setminus\{x_k\}}, Y, Z)$ as follows,

$$\begin{aligned} p(x_k|X_{\setminus\{x_k\}}, Y, Z) &\propto p(Z_{\setminus\{z_k\}}|X_{\setminus\{x_k\}}, Y)p(z_k|x_k,Y)p(x_k|X_{\partial k},Y) \\ &\propto p(z_k|x_k,Y)p(x_k|X_{\partial k},Y) \propto e^{-V_1(z_k|x_k,Y) - \sum_i \alpha_{ik} V_2(x_i,x_k|Y)} \end{aligned}$$

Where $\partial k$ represents the neighbors of node $k$, $\alpha_{ik}$ is 1 if node i is a neighbor of node k, 0 otherwise. In situations where prior is not clear, we can substitute $V_2(x_i, x_k|Y)$ using $V_2(z_i, z_k|x_i, x_k, Y)$. As a local optimization technique, ICM typically converges very fast which results in very efficient semantic similarity assessment. However, convergence to a global minimum is not guaranteed. It is noted that semantic similarities obtained through ICM are unnormalized.

## 4 Experimental Results

We performed multiple sets of experiments to demonstrate the effectiveness of our approach to model individual actions and group activities in a unified way. Though we assume feature points tracking is available across sequences in our experiments, in our approach, trajectories are decomposed into small segments and entities associated with trajectories are not explicitly differentiated, thus, our approach is expected to tolerate inaccurate tracking results. To minimize the affects of translation and scaling, we moved the minimal coordinates ($x_{min}$, $y_{min}$) in each sequence to (0,0), and normalized all coordinate values to [0,1].

### 4.1 Joint Segmentation of Multiple Trajectories

In Sec. 2.2, we perform joint segmentation of multiple trajectories based on the hypothesis that if we cluster trajectories by their major motion components, for individual actions, trajectories associated with the same human body part more often enter the same cluster; for group activities, trajectories associated with collaborative players likely belong to the same group.

To validate our hypothesis for individual action cases, based on the knowledge on human articulations, in Fig. 5a, we suggest five groups of human body landmarks that likely share common simultaneous motions. We perform k-means (k=5) clustering over 3 major components of trajectories from various human actions in the CMU motion capture datasets [19]. As shown in Fig. 5, we found that when landmark points belong to the same group in Fig. 5a, their trajectories more often enter the same cluster.

For group activities in the GA Tech football datasets[1] (sample frames in Fig. 1), as shown in Fig. 6, trajectories in the same cluster tend to exhibit correlated curvature, i.e., similar motion discontinuities in terms of velocity and acceleration, which indicates highly collaborative motions. It is also important to notice from Fig. 6 that the same type of activities tend to generate very similar grouping results, and such property is vital for robust atomic motion segments.

Fig. 7 shows trajectories of three football players from the same group clustered using their major motion components. It can be observed that, very reasonable atomic joint motions can be obtained through multiple trajectory joint segmentation based on the local maxima of the aggregated curvatures.

| |
|---|
| Group 1: {head, left shoulder,right shoulder} |
| Group 2: {left elbow, left wrist, left hand} |
| Group 3: {right elbow, right wrist, right hand} |
| Group 4: {hip, left knee, left ankle, left foot} |
| Group 5: {hip, right knee, right ankle, right foot} |

(a) Human body motion groups

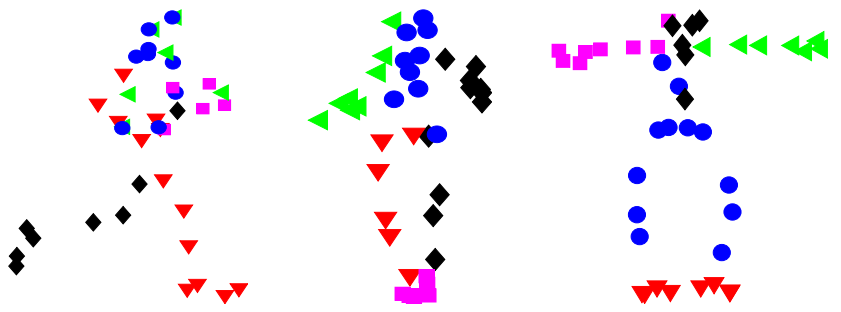

(b) Run (c) Salsa Dance (d) Indian Dance

Fig. 5: Grouping for actions based on common motion. Trajectories at one time instant are shown. The resulting groups are of different shapes and colors.

### 4.2 Group Activity Recognition

We used the GA Tech football dataset to test our approach for modeling complex group activities by classifying football play types. In this dataset, player trajectories in football game videos are available, and football plays are manually labeled into 5 categories. Given the Query-by-Example paradigm assumed, unsupervised classification of each play requires that there exist other similar plays in the rest dataset. Some play types in the given data like *Combo Dropback* actually consists of multiple sub-categories. Since each sub-category consists of only 2 or 3 plays and plays from different sub-categories show sufficiently different patterns, we decide to use three play types of sufficient samples for more reliable performance assessments. These three play types used are, *Drop Back Play*, *Wide Left Run*, and *Wide Right Run*. Each type contains about 8 sequences.

[1] The authors are grateful to GA Tech for providing the datasets.

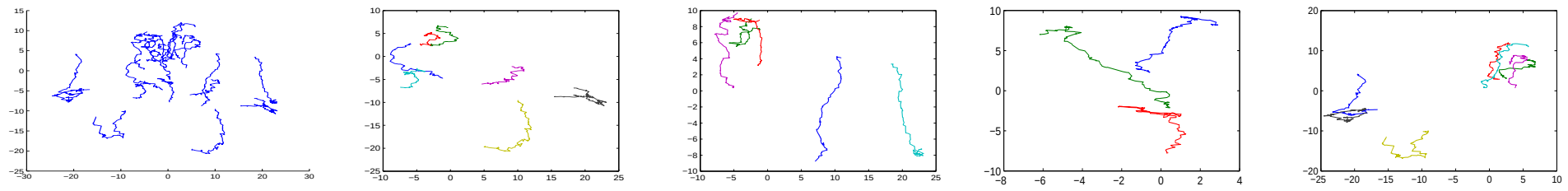

(a) Before grouping (b) group 1 (c) group 2 (d) group 3 (e) group 4

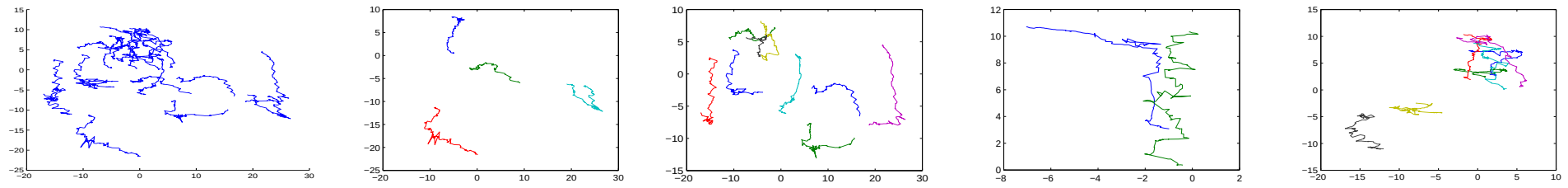

(f) Before grouping (g) group 1 (h) group 2 (i) group 3 (j) group 4

Fig. 6: Trajectories grouping for football *Hitch* play based on common motion. (a)-(e): *Hitch play sequence-1*, (f)-(j): *Hitch play sequence-2*

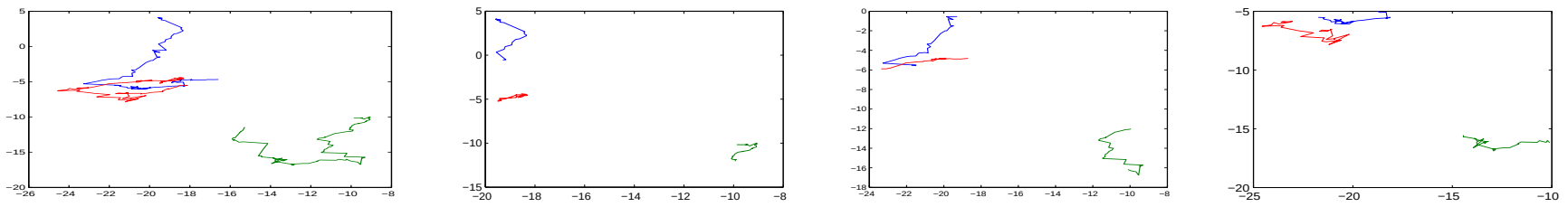

(a) grouped trajectories (b) atomic motion 1 (c) atomic motion 2 (d) atomic motion 3

Fig. 7: Joint segmentation of multiple trajectories to obtain atomic motions

To enable performance comparison, we follow the experimental setups used in [6]. We generate more play samples by applying view transformations with 8 typical views selected from existing plays. We adopt the leave-one-out method. For each football play, we compute its semantic similarity to the rest, and classify it using the 2-NN rule. The classification results are shown in Table 1a.

The results of our approach in classifying group activity are comparable to the one reported in [6], i.e., *DBP 0.77, WLR 0.96, WRR 0.97*. In [6], average recognition rate is also reported for SVM as 0.69. However, it is important to realize that these approaches require learning and make a strong assumption that the correspondence between entities detected in video with those in models are known. To the best of our knowledge, we are not aware of classification schemes which are unsupervised and require no assumption that the order of features are consistent across datasets for fair performance comparison.

### 4.3 Unifying Action and Activity Recognition

To demonstrate that the same model can be directly applied to both individual actions and group activities in a unified way, we add to the football data two complex individual actions, *Salsa Dance* and *Indian Dance* from the motion capture dataset. We have 15 motion sequences for each individual action. Each motion sequence consists of 2D projected trajectories of 41 body landmark points on the human body. To obtain a reliable measure, we applied the same view transforms to action sequences. For each sequence in the merged dataset, which contains a single or multiple persons, we compute its semantic similarity to the rest sequences, and classify it using the 2-NN rule. As shown in Table 1b, our approach can be applied regardless of the number of persons in a sequence.

| | DBP | WLR | WRR |
|---|---|---|---|
| DBP | 0.78 | 0.12 | 0.10 |
| WLR | 0.05 | 0.87 | 0.08 |
| WRR | 0.08 | 0.06 | 0.86 |

(a) Football play

| | DBP | WLR | WRR | Salsa | Indian |
|---|---|---|---|---|---|
| DBP | 0.78 | 0.12 | 0.10 | 0 | 0 |
| WLR | 0.05 | 0.85 | 0.08 | 0.02 | 0 |
| WRR | 0.08 | 0.06 | 0.86 | 0 | 0 |
| Salsa | 0 | 0 | 0 | 0.96 | 0.04 |
| Indian | 0 | 0 | 0 | 0.24 | 0.76 |

(b) Merged football play and individual dance

Table 1: Confusion matrix of football play and dance recognition: Drop Back Play (DBP), Wide Left Run (WLR), Wide Right Run (WRR), Salsa, Indian dance

## 5 Conclusions

This paper presented a unified approach to handle individual actions and group activities. The proposed approach can be used to recognize or retrieve human activity video sequences regardless of the number of persons involved. Our experiments demonstrated the effectiveness of our approach, given it was applied in an unsupervised manner and required no prior on entity correspondences across different sequences. Though we already have some promising initial results, the following issues need to be investigated: 1.) applying the proposed approach in an incremental learning manner, 2.) supporting view invariance without sacrificing the discriminating ability.